\documentclass[12pt, dvipdfmx]{article}
\usepackage{amsmath,amssymb}
\usepackage{times}
\usepackage[dvipdfmx]{graphicx}
\usepackage{multirow}
\usepackage{rotating}
\usepackage{bbm}
\usepackage{latexsym}
\usepackage{parskip,comment}
\usepackage[normalem]{ulem}
\usepackage{braket}
\usepackage{here}
\usepackage{algpseudocode,algorithm}
\usepackage[natbibapa]{apacite}
\usepackage[subrefformat=parens]{subcaption}

\def\0{{\boldsymbol 0}}

\def\k{{\boldsymbol k}}

\def\v{{\boldsymbol v}}

\def\h{{\boldsymbol h}}

\def\b{{\boldsymbol b}}
\def\c{{\boldsymbol c}}

\def\bxi{{\boldsymbol \xi}}
\def\l<{\left<} \def\r>{\right>}
\def\wt{\widetilde}
\def\tra{{}^t\!} 

\def\changed#1{{\bf\color[rgb]{1,0,0} #1}}

\newcommand{\com}[1]{{\sf\color[rgb]{0,0,1}{#1}}}
\usepackage[dvips]{color} % For blue in-text comments and additions
\newcommand{\DKL}[2]{D_{\rm KL}({#1}||{#2})}

\newcommand{\argmin}{\mathop{\rm arg~min}\limits}
\newcommand{\captionfonts}{\normalsize}

\makeatletter  
\long\def\@makecaption#1#2{%
  \vskip\abovecaptionskip
  \sbox\@tempboxa{{\captionfonts #1: #2}}%
  \ifdim \wd\@tempboxa >\hsize
    {\captionfonts #1: #2\par}
  \else
    \hbox to\hsize{\hfil\box\@tempboxa\hfil}%
  \fi
  \vskip\belowcaptionskip}
\makeatother   
\begin{document}
\hspace{13.9cm}1

\ \vspace{20mm}\\

{\LARGE Decreasing the size of the Restricted Boltzmann machine}

\ \\
{\bf \large Yohei Saito}\\
{yoheis@sat.t.u-tokyo.ac.jp \\
Institute of Industrial Science, The University of Tokyo, 4-6-1, Komaba, \\
Meguro-ku, Tokyo 153-8505 Japan}\\
{\bf \large Takuya Kato}\\
{takuya.kato.origami@gmail.com \\
Graduate School of Information Science and Technology, \\
Department of Mathematical informatics, The University of Tokyo, 7-3-1, Hongo, \\
Bunkyo-ku, Tokyo 113-8654, Japan}\\
%

%\ \\[-2mm]
%%%% {\bf Keywords:} Manuscript, journal, instructions

\thispagestyle{empty}
\markboth{}{NC instructions}
\ \vspace{-0mm}\\
%
%Abstract
\newpage
\begin{center} {\bf Abstract} \end{center}
In this paper, 
we propose a method to decrease the number of hidden units of the restricted Boltzmann machine 
while avoiding a decrease in the performance quantified by the Kullback-Leibler divergence. 
Our algorithm is then demonstrated by numerical simulations. 
%%%%%%%%%%%

 \section{Introduction}
The improvement of computer performance enables 
utilization of the exceedingly high representational powers of neural networks. 
Deep neural networks have been applied to various types of data, e.g. images, speech, and natural language, 
and have achieved great success 
(\cite{bengio2013representation, he2016deep, vaswani2017attention, goodfellow2014generative, oord2016wavenet}) 
both in discrimination and generation tasks. 
%, which leads to great success of deep neural networks both in discrimination and generation of data, 
%e.g. images, speech, and natural language 
%(\cite{bengio2013representation, he2016deep, vaswani2017attention, goodfellow2014generative, oord2016wavenet}). 
To increase performance, 
which stems from the hierarchical structures of neural networks (\cite{hestness2017deep}), 
network size becomes larger, and computational burdens increase. 
Thus, demands for decreasing the network size are growing. 
In particular, various methods were proposed for compressing the sizes of discriminative models 
(\cite{han2015learning, guo2016dynamic, cheng2017survey}). 
However, compression of generative models (\cite{berglund2015measuring}) has scarcely been discussed.

Discriminative models provide the probabilities that into which class the given data are classified 
(\cite{christopher2016pattern}), 
% \com{(class probability という表現は使わなくないようだが、どうするか)}
and in most cases, their learning requires a supervisor, namely, a dataset with classification labels attached by humans. 
Thus, outputs of discriminative models can be intuitively interpreted by humans. 
However, some data are difficult for humans to properly classify. 
Even if possible, hand-labeling tasks are a troublesome labor. 
In such cases, generative models with unsupervised learning are effective, 
since they automatically find the data structure without hand-labels 
by learning the joint probabilities of data and classes. 
%\sout{However, classification done by generative models may not be easily interpreted by humans. }
%\com{(わざわざ言う必要がない気がしてきた)}
Therefore, it is expected that 
the compression of generative models with unsupervised learning will be required in the future. 
Furthermore, if the system's performance can be preserved during compression, 
then the network size can be decreased while it is in use. 
To approximately maintain performance throughout compression, 
we consider removing the part of the system after decreasing its contribution to the overall performance. 
Our approach differs from the procedures in previous studies 
(\cite{han2015learning, guo2016dynamic, cheng2017survey, berglund2015measuring}) 
that retrain systems after removing a part that contributes little to their performance.

\begin{figure}[t]
% \begin{figure}[t]
\centering
% \begin{center}
  \includegraphics[width=90mm]{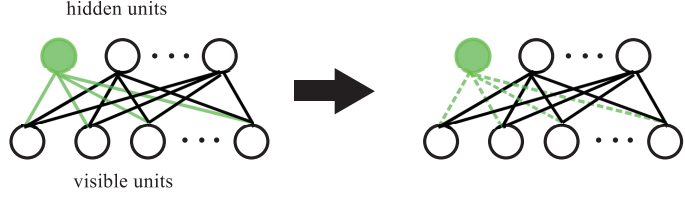}
% \end{center}
 \caption
{The graphical model of the RBM is shown. 
While approximately preserving the KLD, 
the target hidden unit and its edges (green) are removed from the main body of the RBM 
(from the left to right panel). \vspace{0.7cm}
% We notice that our proposal method does not cut edges of the target node, as explained in Sec.3.
}
 \label{fig:RBM}
\end{figure}

In this paper, we deal with the restricted Boltzmann machine (RBM) 
(\cite{smolensky1986information, fischer2012introduction}). 
The RBM is one of the most important generative models with unsupervised learning, 
from the viewpoints of not only machine learning history (\cite{bengio2013representation}) 
but also its wide applications, 
e.g., generation of new samples, classification of data (\cite{larochelle2008classification}), 
feature extraction (\cite{hinton2006reducing}), 
pretraining of deep neural networks (\cite{hinton2006reducing, hinton2006fast, salakhutdinov2010efficient}), 
and solving many-body problems in physics (\cite{carleo2017solving, tubiana2017emergence}). 
The RBM consists of visible units that represent 
% \sout{components of }
observables, e.g., pixels of images, 
and hidden units that express correlations between visible units. 
An objective of the RBM is to generate plausible data 
by imitating the probability distribution from which true data are sampled. 
% that is, to obtain plausible observables such as images. 
In this case, 
the performance of the RBM is quantified by the difference between the probability 
distribution of data and that of visible variables of the RBM, 
and it can be expressed by the Kullback-Leibler divergence (KLD). 
% Since the number of hidden units determines representational power of the RBM, 
% which is related to upper bound of performance, 
% they can be regarded as resources on which we focus. 
The RBM can exactly reproduce any probability distribution of binary data 
% can express arbitrary probability, 
if it has a sufficient number of hidden units (\cite{le2008representational}). 
However, a smaller number of hidden units may be enough to capture the structure of the data. 
Therefore, in this paper, 
we aim to practically decrease the number of hidden units 
while avoiding an increase in the KLD between the model and data distributions (Figure~\ref{fig:RBM}). 
% \com{(in each step??)} 
% In this paper, we propose a method to change network structure of the BM 
% and evaluate the variation of the KLD caused by this method. 
% Then, we examine our method by some numerical simulations. 

% Then, we examine our method by numerical simulations. 
% Owing to the high computational cost of the BM, 
% we employ the deep Boltzmann machine (DBM), which is obtained by amputating some of edges of the BM (Fig.~\ref{fig:BM_cut}). 

% For practical applications, 
% a part of the BM called restricted Boltzmann machine (RBM), which has connections solely between visible and hidden nodes, 
% has been used as generation or discrimination of observables, or pre-training of the deep neural networks. 
% As an extensive model of RBM, undirected graph with many hidden layers called the deep Boltzmann machine (DBM) is proposed. 

% It is meaningful to consider flexible size BM for applications in neural networks. 

The outline of this paper is as follows. 
In section 2, we give a brief review of the RBM. 
In section 3, we evaluate the deviation of the KLD associated with node removal 
and propose a method that decreases the number of hidden units while avoiding an increase in the KLD. 
Numerical simulations are demonstrated in section 4, 
% and give an explanation of the results. 
and we summarize this paper in section 5. 
The details of calculations are shown in Appendices.

\begin{comment}
\com{(Move to Sec. 2.)}
It is easy to increase network size except for over fitting problems, 
since adding new nodes which do not contribute system's functions at all does not cause variation of system's functions. 
Hence, if a whole system slowly relearns, network size increases while approximately keeping performances. 
In contrast, decrease of network size inevitably causes variation of performances in most of cases. 
It can be considered that in order to reduce this variation, 
one makes a unit to be removed transfer its role into other units of a system 
before removing it. 
Thus, one should estimate the variation of performance which is caused by transfer and removing steps. 
\end{comment}

 \section{Brief introduction of the RBM}
In this section, we briefly review the RBM, 
which is a Markov random field that consists of visible units, 
% which represent components of observables, e.g., pixels of images, 
$\tra\v=(v_1, \ldots, v_M)\in \{ 0,1\}^M\,$, 
and hidden units, 
% which express correlation between visible units, 
$\tra\h=(h_1,\ldots, h_N)\in \{ 0,1\}^N\,$. 
% (Fig.~\ref{fig:RBM}). 
The joint probability that a configuration $(\v, \h)$ is realized, $p(\v,\h)\,$, 
is given by the energy function, $E(\v,\h)\,$, as follows: 
 \begin{eqnarray}
  E(\v,\h)
    &=& -\tra\b \, \v - \tra\c \, \h - \tra\v \, W \, \h \nonumber \\
    &=& - \sum_{i=1}^M b_i \, v_i - \sum_{j=1}^N c_j \, h_j  
     - \sum_{i=1}^M \sum_{j=1}^N v_i \, w_{ij}\, h_j \, , \label{RBM_Energy} \\[0.3cm]
  p(\v,\h) &=& \frac{{\rm e}^{-E(\v,\h)}} {\sum_{\v',\h'} {\rm e}^{-E(\v',\h')}} \, , 
 \end{eqnarray}
where $\tra \b=(b_1,\ldots, b_M) \in\mathbb R^M$ and 
$\tra \c=(c_1,\ldots, c_N) \in\mathbb R^N$ are the biases of the visible and hidden units, respectively, 
and $W=(w_{ij}) \in\mathbb R^{M \times N}$ is the weight matrix
\footnote{
The RBM whose visible and hidden units take $\tra\v' \in \{ -1,1\}^M$ and $\tra\h' \in \{ -1,1\}^N$ 
can be related to the RBM that takes $\tra\v \in \{ 0,1\}^M$ and $\tra\h \in \{ 0,1\}^N$ 
by changing the parameters, 
$W' = W / 4\,$, $b'_i = b_i / 2 + \sum_j w_{ij} / 4$ and $c'_j = c_j / 2 + \sum_i w_{ij} / 4\,$, 
where $\b'\,$, $\c'$ and $W'$ are the biases and weight matrix of the RBM whose nodes take $\{ -1,1 \}\,$. 
}
. 
Below, we abbreviate all of the RBM parameters, $\b\,$, $\c\,$, and $W\,$, as $\bxi\,$. 
%of the edges between nodes in visible and hidden layer. 
% a node $a$ in $\mu$ layer, $h^\mu_a\,$, and a node $b$ in $\nu$ layer, $h^\nu_b\,$. 
% In this paper, $\xi_i$ denotes one of the parameters of DBM, e.g., bias or weight. 

%The sum of some function $f(\v,\h)$ over $\v$ and $\h$ means the sum over all of the configurations of $\v$ and $\h\, $,
% \begin{eqnarray}
%  && \sum_{\v,\h} f(\v, \h) \nonumber \\ 
%  && = \sum_{v_1=0,1} \cdots \sum_{v_M=0,1} \sum_{h_1=0,1} \cdots \sum_{h_N=0,1} f(\v, \h) \, .
% \end{eqnarray}

% , 
% that is, for some function $F(\h^\mu)\,$, 
%  \begin{eqnarray}
%   && \sum_{\h^\mu} F(\h^\mu) \nonumber \\
%   && \equiv \sum_{h^\mu_1=0,1}\, \sum_{h^\mu_2=0,1}\cdots\sum_{h^\mu_{N_\mu}=0,1} F(h^\mu_1,\ldots, h^\mu_{N_\mu}) \, . 
%  \end{eqnarray}

%The visible probability generated by BM, $p(\v)\,$, is obtained 
%by marginalizing $p(\v,\h)$ with respect to all of the configurations of the hidden nodes, 
% \begin{eqnarray}
%  p(\v) = \sum_\h p(\v,\h) \, .
% \end{eqnarray}

By properly tuning $\bxi\,$, 
%based on samples of observables, 
the probability distribution of the visible variables, $p(\v) = \sum_\h p(\v,\h)\,$, 
can approximate the unknown probability distribution that generates real data, $q(\v)\,$. 
% Since an objective of the RBM is to generate plausible observables from $p(\v)\,$, 
%A goal of the RBM is to produce plausible observables from $p(\v) = \sum_\h p(\v,\h)$ 
%that approximates the probability, $q(\v)\,$, which generates true observables. 
% by tuning the parameters of the BM, namely, biases and weights. 
The performance of the RBM can be measured by the KLD of $p(\v)$ from $q(\v)\,$, 
% \com{(from $p(\v)$ to $q(\v)$?)}
 \begin{eqnarray}
  \DKL{q}{p} = \sum_\v q(\v)\, \ln\frac{q(\v)}{p(\v)} \, .
  \label{def:KLD}
 \end{eqnarray} 
Hence, learning of the RBM is performed by updating the RBM parameters $\bxi$ so as to decrease the KLD. 
%namely, $\b\,$, $\c$ and $W\,$, so as to decrease the KLD. 
The gradient descent method is often employed to decrease the KLD as 
 \begin{eqnarray}
  \bxi^{s+1} = \bxi^s - \lambda \, \nabla_\bxi \DKL{q}{p}|_{\bxi = \bxi^s} \, , 
  \label{gradient-descent}
 \end{eqnarray}
%or 
% \begin{eqnarray}
%  \xi_i(t+1) = \xi_i(t) -\alpha\, \frac{\partial \DKL{q}{p}}{\partial \xi_i(t)} \, ,
%  \label{gradient-descent}
% \end{eqnarray}
where $\bxi^s$ and $\bxi^{s+1}$ denote the RBM parameters at the $s$-th and $(s+1)$-th step of the learning process, respectively. 
A learning rate is represented by $\lambda\, (>0)\,$, 
and $\nabla_\bxi \DKL{q}{p}|_{\bxi=\bxi^s}$ denotes the gradient of the KLD with respect to $\bxi$ at the $s$-th step. 
%and $\xi_i$ represents one of a parameter of the BM, that is, a bias or weight. 
%$\xi_i(t)$ and $\xi_i(t+1)$ denote parameters before and after update, respectively. 
%Each component of the gradient can be obtained from 
% \begin{eqnarray}
%  \frac{\partial \DKL{q}{p}}{\partial \xi_i} 
%   &=& -\sum_{\v,\h} \frac{\partial E(\v,\h)}{\partial \xi_i} \, q(\v)\, p(\h|\v) \nonumber \\
%   && +\sum_{\v,\h} \frac{\partial E(\v,\h)}{\partial \xi_i} \, p(\v,\h) \, ,
%  \label{gradient} 
% \end{eqnarray}
%where $p(\h|\v)$ is the conditional probability, 
% \begin{eqnarray}
%  p(\h|\v) = \frac{p(\v,\h)}{p(\v)} \, .
% \end{eqnarray} 
The gradient with respect to $b_i, c_j,$ and $w_{ij}$ can be written as
 \begin{eqnarray}
  \frac{\partial D}{\partial b_i} 
   % &=& -\sum_\v q(\v) \sum_\h v_i \, p(\v|\h) + \sum_{\v, \h} v_i \, p(\v, \h) \nonumber \\
   &=& -\sum_\v v_i \, q(\v) + \left< v_i \right>_p \, , \label{D_b} \\
  \frac{\partial D}{\partial c_j} 
   % &=& -\sum_\v q(\v) \sum_\h h_j \, p(\v|\h) + \sum_{\v, \h} h_j \, p(\v, \h) \nonumber \\
   &=& -\sum_\v q(\v) \, p(h_j = 1 | \v) + \left< h_j \right>_p \, , \label{D_c} \\
  \frac{\partial D}{\partial w_{ij}} 
   % &=& -\sum_\v q(\v) \sum_\h v_i \, h_j \, p(\v|\h) + \sum_{\v, \h} v_i \, h_j \, p(\v, \h) \nonumber \\
   &=& -\sum_\v v_i \, q(\v)\, p(h_j=1 | \v) + \left< v_i h_j \right>_p \, , \label{D_w}
 \end{eqnarray}
where $\DKL{q}{p}$ is abbreviated as $D$ 
and the expectation value with respect to $p(\v, \h)$ as $\left< \cdot \right>_p\,$. 
The conditional probability, $p(h_j | \v)\,$, is given by 
 \begin{eqnarray}
  p(h_j | \v) = \frac{ {\rm e}^{(c_j + \sum_i v_i w_{ij})\, h_j} }{ 1 + {\rm e}^{c_j + \sum_i v_i w_{ij}} } \, . 
 \end{eqnarray}

If $D$ and $\nabla_\bxi D$ can be obtained, 
then the RBM reaches some local minimum of the KLD through a parameter update. 
However, neither of them can be calculated, 
since they not only contain the unknown probability $q(\v)$ 
but also the sum with respect to the large state space of the RBM.  
Thus, in Eq.~(\ref{D_b}), Eq.~(\ref{D_c}), and Eq.~(\ref{D_w}), 
One approximates $q(\v)$ by empirical distribution, or more practically, 
mini-batch, which are samples from the empirical distribution. 
One also evaluates the expectation values with respect to $p(\v,\h)$, which are computationally expensive, 
by using the realizations obtained from Gibbs sampling, 
e.g. contrastive divergence (CD) (\cite{hinton2002training}), 
persistent CD (PCD) (\cite{tieleman2008training}), fast PCD (\cite{tieleman2009using}), 
and block Gibbs sampling with tempered transition (\cite{salakhutdinov2009learning}) 
or with parallel tempering (\cite{cho2010parallel, desjardins2010tempered}). 
Block Gibbs sampling in the RBM effectively updates the configuration, $(\v, \h)\,$, 
by repeatedly using the conditional probabilities, 
 \begin{eqnarray}
  p(\h|\v) &=& \prod_j p(h_j | \v) = \prod _j \frac{ {\rm e}^{(c_j + \sum_i v_i w_{ij})\, h_j} }{ 1 + {\rm e}^{c_j + \sum_i v_i w_{ij}} } \, , 
  \label{propup}\\
  p(\v|\h) &=& \prod_i p(v_i | \h) = \prod _i \frac{ {\rm e}^{(b_i + \sum_j h_j w_{ij})\, v_i} }{ 1 + {\rm e}^{b_i + \sum_j h_j w_{ij}} } \, ,
  \label{propdown}
 \end{eqnarray}
as transition matrices. 
In many cases, CD and PCD employ only a few block Gibbs sampling steps. 
In addition to $\nabla_\bxi D\,$, the KLD, which represents the performance of the RBM, is also intractable. 
Therefore, in order to monitor the learning progress, 
a different quantity is employed which can be considered to correlate to the KLD to a certain degree, 
e.g. the reconstruction error (\cite{bengio2007greedy, taylor2007modeling, hinton2012practical}), 
the product of the two probabilities ratio (\cite{buchaca2013stopping}), 
and the likelihood of a validation set obtained by tracking the partition function (\cite{desjardins2011tracking}). 

% and $K_i$ is defined by 
%  \begin{eqnarray}
%   K_i \equiv \frac{\partial E(\h^0,\h)}{\partial\xi_i} \, .
%  \end{eqnarray}
% Due to the wide state space of $p(\h^0,\ldots,\h^D)$ and $p(\h^,\ldots,\h^D|\h^0)\,$, 
% calculation of Eq.~(\ref{gradient}) is not realistic, and hence, Gibbs sampling is employed to estimate $\nabla_\bxi \DKL{q}{p}\,$. 
% Thus, Eq.~(\ref{gradient}) is stochastic value in practice. 

  \section{Removal of hidden units}
   \subsection{Removal cost and its gradient}
The goal of this paper is not to propose a new method for optimization of the KLD, 
but to decrease the number of hidden units while avoiding an increase in the KLD. 
%but to change the number of the hidden units while keeping $\DKL{q}{p}\,$. 
%In the first subsection, we evaluate the deviation of the KLD when a hidden unit is removed, 
%and propose how to safely remove a hidden node. 
%In the second subsection, we demonstrate numerical simulations. 
%	In this section, we mainly discuss how to decrease the number of hidden units, 
%	and shortly comment on addition of hidden units at the end of this Section. 
%keeping performance. 
%Therefore, in this section, we evaluate the change of the KLD as $N$ changes 
%and propose a method to change $N$ while preserving the KLD. 
%When the RBM shows adequate performance, one changes the number of hidden nodes. 
%We denote $\bxi^0$ as parameters of the RBM after training procedure, 
%and represent $p_0(\v,\h)$ and $D_0$ as the probability and the KLD at $\bxi = \bxi^0\,$: 
% \begin{eqnarray}
%  D_0 &=& \sum_\v q(\v) \, \ln \frac{q(\v)}{p_0(\v)} \, , \\
%  p_0(\v, \h) &=& \frac{{\rm e}^{-E_0(\v,\h)} }{ \sum_{\v', \h'} {\rm e}^{-E_0(\v',\h')} } \, , \\
%  E_0(\v, \h) &=& E(\v, \h) |_{\bxi = \bxi^0} \, .
% \end{eqnarray} 
Suppose an RBM achieves, if not optimal, 
sufficient performance after the learning process at a fixed number of hidden units, $N\,$. 
Next, we remove the $k$-th hidden unit of the RBM so as not to increase the KLD. 
% Then, one attempts to decrease calculation cost by decreasing $N$ while continuing to use the RBM. 
% Suppose one removes $k$-th hidden node of the RBM. 
In order to compare the performances of two RBMs whose $k$-th hidden unit does or does not exist, 
% consider the deviation of the KLD when a hidden node $h_k$ is removed. 
we introduce $\h_{\backslash k}$ as a configuration of hidden units except for $h_k\,$, 
$\tra \h_{\backslash k} = (h_1,\ldots, h_{k-1}, h_{k+1}, \ldots, h_N)\,$. 
The energy function and the probability distribution of the RBM after removal are given by 
 \begin{eqnarray}
  E_{\backslash k} (\v, \h_{\backslash k}) 
    &=& - \sum_i b_i \, v_i - \sum_{j \neq k} c_j \, h_j  
     - \sum_i \sum_{j \neq k} v_i \, w_{ij}\, h_j \nonumber \\ 
    &=& E(\v, \h )|_{h_k=0} \, , \label{RBM_Energy_k} \\[0.3cm]
  p_{\backslash k}(\v, \h_{\backslash k})
    &=& \frac{ {\rm e}^{-E_{\backslash k}(\v, \h_{\backslash k}) } }
                  { \sum_{ \v', \h'_{\backslash k} } {\rm e}^{-E_{\backslash k}(\v', \h'_{\backslash k}) } } \, . 
    \label{p-k}
 \end{eqnarray}
Then, we define a removal cost, $C_k\, $, as the difference of the KLD before and after removing the $k$-th hidden unit: 
 \begin{eqnarray}
  C_k &\equiv& \DKL{q}{p_{\backslash k}} - \DKL{q}{p} \nonumber \\ 
   &=& \sum_\v q(\v) \, \ln \frac{q(\v)}{p_{\backslash k} (\v)} - \sum_\v q(\v) \, \ln \frac{q(\v)}{p(\v)} \nonumber \\[0.3cm]
   &=& -\sum_\v q(\v) \, \ln p(h_k = 0 | \v) + \ln p(h_k = 0) \, .
%   &=& -\sum_\v q(\v) \, \ln \left[ 1 - p(h_k = 1 | \v) \right] \nonumber \\ 
%   && + \ln \left[ 1 - p(h_k = 1) \right] \, . 
  \label{remove_cost}
 \end{eqnarray}
The details of the calculation and removal cost for several hidden units are shown in Appendix A. 
Thus, if $C_k$ satisfies $C_k\leq 0\,$, 
then the $k$-th hidden unit can be removed without increasing the KLD.

In most cases, however, there are no hidden units with non-positive removal costs. 
Thus, before removing a hidden unit, we first decrease its removal cost without increasing the KLD
\footnote{
As explained in Appendix A, minimizing the size of the RBM 
%\sout{, which we do not discuss in this paper, can be expressed as follows. 
%Find a set of parameters $\bxi$ that maximize the number of hidden units whose removal costs are zero 
%subject to  $D = D_0\,$, where $D_0$ is the KLD to be preserved. 
%Since this problem }
is a difficult problem. 
Thus, in this paper, hidden units are removed individually in a greedy fashion. 
%If one attempts to remove a hidden node while keeping the KLD, 
%one should find a set of parameters $\bxi$ which satisfies $C_k = 0$ and
%More practically, one minimizes $(C_k)^2$ subject to $D=D_0\,$, 
%and when $(C_k)^2 = 0$ is achieved, $k$-th hidden node can be removed without changing the KLD. 
%However, this constrained optimization problem is difficult to deal with, 
%since both $C_k$ and $D$ cannot be evaluated accurately, 
%and use of Gibbs sampling causes inaccurate numerical results and long calculation time. 
}. 
% $C_k$ should be decreased, and the KLD should be preserved. 
% Therefore, in order to avoid increasing the KLD, we update $\bxi$ so as to decrease both $C_k$ and $D\,$
%Thus, we minimize $(C_k)^2$ subject to $\Delta D = D - D_0 = 0\,$. 
%Although such a constrained optimization problem can be deal with Lagrange multiplier, 
%it is numerically hard to solve nonlinear equations to find a stationary point of Lagrange multipliers. 
%Instead, we minimize cost function, 
% \begin{eqnarray}
%  L_k = \frac{1}{2} \, (C_k)^2 + \frac{\lambda}{2} \, (\Delta D)^2 \, , 
% \end{eqnarray}
%by using gradient descent while increasing $\lambda\,$. 
%That is, we first minimize $L_k$ at fixed $\lambda\,$, 
%and then, we increase $\lambda$ and minimize $L_k$ again. 
%In this way, we attempt to minimize $(C_k)^2$ subject to $\Delta D=0\,$. 
%Thus, we update the parameters so as to decrease both the KLD and $C_k$ at the first order of ${\cal O}(\Delta\bxi)\, $. 
%When $\bxi^{s+1} = \bxi^s + \Delta\bxi\,$, $C_k$ and $D$ at $(s+1)$-step can be written as 
% \begin{eqnarray}
%  C_k^{s+1} &=& C_k^s + \tra \Delta\bxi \, \nabla_\bxi \, C_k^s + {\cal O}(\Delta\bxi^2) \, , \\
%  D^{s+1} &=& D^s + \tra \Delta\bxi \, \nabla_\bxi \, D^s + {\cal O}(\Delta\bxi^2) \, , 
% \end{eqnarray}
For this purpose, we naively determine the parameter update at the $s$-th step in a removal process, $\Delta\bxi^s\,$, 
so that both $C_k$ and the KLD decrease at ${\cal O}(|\Delta\bxi^s|)$ (see Appendix B): 
 \begin{eqnarray}
  \Delta\xi_i^s 
   &=& - \nu \cdot \theta \left( \frac{\partial D}{\partial \xi_i} \, \frac{\partial C_k}{\partial \xi_i} \right)  
        \cdot \frac{\partial D}{\partial \xi_i} \, \bigg|_{\bxi = \bxi^s} \, , \label{update_2} \\
  \theta(x)
   &=& 
   \begin{cases}
    1 & (x \geq 0) \\
    0 & (x < 0)
   \end{cases} \, ,
 \end{eqnarray}
where $\nu \, (>0)$ is the parameter change rate, and $\theta(x)$ is the step function. 
Evaluation of $\nabla_\bxi D$ can be performed using Eq.~(\ref{D_b}), Eq.~(\ref{D_c}), and Eq.~(\ref{D_w}), 
and $\nabla_\bxi C_k$ can be written as 
% \begin{eqnarray}
%  \nabla_\bxi L_k = C_k \, \nabla_\bxi C_k + \lambda \, \Delta D \, \nabla_\bxi D \, ,  
% \end{eqnarray}
%where $\nabla_\bxi D$ is given by Eqs.~(\ref{D_b}), (\ref{D_c}) and (\ref{D_w}). 
%The first derivatives of $C_k$ are given by 
 \begin{eqnarray}
  \frac{\partial C_k}{\partial b_i}
   &=& \left< v_i \right>_{\bar p} - \left< v_i \right>_p \, , \\
  \frac{\partial C_k}{\partial c_j}
   &=& \sum_\v q(\v) \, p(h_k=1 | \v) \, \delta_{kj} 
    + \left< h_j \right>_{\bar p} - \left< h_j \right>_p \, , \\
  \frac{\partial C_k}{\partial w_{ij}}
   &=& \sum_\v q(\v) \, v_i \, p(h_k=1 | \v) \, \delta_{k j} 
    + \left< v_i h_j \right>_{\bar p} - \left< v_i h_j \right>_p \, ,
 \end{eqnarray}
where $\delta_{kj}$ is the Kronecker delta, 
and $\left< \cdot \right>_p$ and $\left< \cdot \right>_{\bar p}$ denote expectation values 
with respect to $p(\v, \h)$ and $\bar p \equiv p(\v, \h_{\backslash k} | h_k = 0)\,$, respectively. 
If $C_k \leq 0$ is satisfied after parameter updates, 
then the $k$-th hidden unit can be removed without increasing the KLD. 
When all of the RBM parameters satisfy $\partial D/\partial\xi_i \cdot \partial C_k/\partial\xi_i < 0\,$, 
then $C_k$ cannot decrease without increasing the KLD, 
and the parameter update is stopped ($\Delta\bxi = \0$)
\footnote{
For $\nabla_\bxi D = \0\,$, which seldom occurs in numerical simulations, 
we employ higher-order derivatives of $D$ 
and seek a direction along which both $C_k$ and $D$ decrease. 
By restricting the number of parameters to be updated, 
one can alleviate computational cost caused by a large number of the elements of higher-order derivatives. 
}
. 

Note two properties of $C_k\,$. 
First, $-C_k$ can be interpreted as an additional cost of a new node. 
Thus, it may be employed when new nodes are added into an RBM whose performance is insufficient. 
% we consider a method for adding a new hidden unit without changing the KLD, 
% , $C_k = 0\,$, 
% when performance of the RBM is insufficient. 
	%From Eq.~(\ref{remove_cost}), one finds that 
	%if all of the weights of edges of a new node are $0\,$, 
	%$C_k$ becomes $0$ (see Appendix B). 
	%Therefore, addition of such a new hidden unit does not harm the system's performance, 
	%although it is known that learning of additional new units takes long time. 
	%\com{(Cite infinite RBM?)}
Secondly, Eq.~(\ref{remove_cost}) can be applied to the Boltzmann machine (BM) (\cite{ackley1987learning}), 
which is expressed as a complete graph consisting of visible and hidden units, 
and a special case of the BM called the deep Boltzmann machine (DBM) (\cite{salakhutdinov2009deep}), 
which has hierarchical hidden layers with neighboring interlayer connections. 
However, in these cases, calculation of the conditional probability, $p(h_k=0|\v)\,$, 
and gradients with respect to the model parameters are computationally expensive compared to the RBM.

   \subsection{Practical removal procedure}
The removal process proposed in the previous subsection preserves the performance 
when $C_k\,$, $\nabla_\bxi C_k\,$, and $\nabla_\bxi D$ can be accurately evaluated. 
However, in most cases, $C_k$ and $\nabla_\bxi C_k$ are approximated using Gibbs sampling, as with $\nabla_\bxi D\,$. 
Thus, in order to reflect the variances of Gibbs sampling, 
we change both the parameter update rule and removal condition, 
Eq.~(\ref{update_2}) and Eq.~(\ref{remove_cost}), into more effective forms.

First, we modify the parameter update rule, Eq.~(\ref{update_2}), which may increase $D$ due to two reasons. 
The first is the inaccuracy of Gibbs sampling, 
and the second is the contribution from higher-order derivative terms of ${\cal O}(|\Delta\bxi|^2)\,$. 
These problems also arise in the learning process. 
However, even if $D$ increases, it can decrease again through the update rule, Eq.~(\ref{gradient-descent}). 
Since the difference between Eq.~(\ref{gradient-descent}) and Eq.~(\ref{update_2}) is 
solely the existence of the step function, 
similar behavior is expected in the removal process. 
%It is expected that the update rule, Eq.~(\ref{update_2}), 
%eventually compensate the occasional increase of $D$ caused by the above-listed two reasons, 
%since the update rule is designed to decrease $D$ at ${\cal O}(|\Delta\bxi|)\,$. 
Unfortunately, Eq.~(\ref{update_2}) frequently increases $D$ due to the following. 
%When $D$ increases, update rule of removing process is expected to decrease $D\,$, 
%as with that of training process. 
%However, we cannot expect decrease of $D$ from the following reason. 
Since the removal cost is defined as the change in the KLD through node removal, 
it can be interpreted as the contribution of the node to the performance. 
Hence, when the performance increases, removal costs are expected to increase. 
This means that in the RBM parameter space, 
there are few directions along which both $D$ and $C_k$ decrease. 
However, since the step function in Eq.~(\ref{update_2}) allows the parameter update solely along these few directions, 
there are few opportunities to decrease $D\,$. 
% once it increases, which may result successive increase of $D\,$. 
%\sout{
%In learning process, even if the KLD increases 
%owing to inaccuracy of Gibbs sampling estimates or contribution from higher order terms, ${\cal O}(|\bxi^s|^2)\,$, 
%it can decrease again following Eq.~(\ref{gradient-descent}). 
%However, in parameter update rule given by Eq.~(\ref{update_2}), 
%the step function prevents parameter updates along the direction given by $\nabla_\bxi D\,$, 
%when $C_k$ increases along this direction. 
%Thus, Eq.~(\ref{update_2}) may result successive decrease of $D\,$. 
%In order to avoid this successive increase, we require two properties to modified update rule. 
%One is that it can update parameters along the direction of $\nabla_\bxi D$ in principle, 
%being biased along the direction of $\nabla_\bxi C_k\,$. 
%The other is that it returns to Eq.~(\ref{update_2}), when Gibbs sampling estimates are exactly obtained. 
%}
%\sout{
%In order to update parameters while approximately keeping the KLD, 
%we partially admit increase of $C_k\,$. 
%}
%\sout{
%Since the step function in Eq.~(\ref{update_2}) prevents decrease of $C_k$ with increase of $D\,$, 
%we may no longer decrease $D$ after accidentally increase of $D\,$, 
%which may cause successive increase of $D\,$. 
%Hence, taking into account inaccuracy of Gibbs sampling, 
%}
Therefore, once $D$ increases, it rarely decreases by Eq.~(\ref{update_2}). 
As a result, a successive increase of $D$ occurs. 
%more frequently 
%than decrease of $D$ at any steps of the removal process, 
%which leads to average increase of $D\,$. 
In order to maintain the performance, we probabilistically accept updates which increase $C_k\,$. 
That is, we change the step function in Eq.~(\ref{update_2}), which gives either $0$ or $1$ deterministically, 
into a random variable, $z_i \in \{0, \, 1\}\,$. 
Next, we determine the probability that $z_i$ takes $1\,$, that is, the acceptance probability of updates. 
The modified update rule is required to return to Eq.~(\ref{update_2}) 
when Gibbs sampling estimates are exactly obtained. 
For this purpose, we employ the ratio of the mean to the standard deviation and determine 
% \sout{which is the inverse of the relative standard deviation}. 
the modified update rule by 
 \begin{eqnarray}
  \overline{\Delta\xi_i^s} 
   &=& - \nu \, z_i  \, \overline{\partial_i D} \, |_{\bxi = \bxi^s} \, , \label{update_2mod} \\[0.3cm]
   p(z_i=1) &=& {\rm sig}\left( \frac{\sqrt{S} \cdot \overline{\partial_i D}} {\overline{\sigma_{D,i}}} \cdot  
                     \frac{\sqrt{S} \cdot \overline{\partial_i C_k}} {\overline{\sigma_{C,i}}} \right) \, , \\[0.3cm] 
  {\rm sig}(x) &=& \frac{{\rm e}^x}{1+{\rm e}^x} \, ,  
 \end{eqnarray}
where $S$ is the number of Gibbs samples, 
and $\overline{\partial_i D}$ and $\overline{\partial_i C_k}$ represent 
sample means of $\partial D / \partial \xi_i$ and $\partial C_k / \partial \xi_i$, respectively. 
The unbiased standard deviations of $\partial D / \partial \xi_i$ and $\partial C_k / \partial \xi_i$ 
are denoted by $\overline{\sigma_{D,i}}$ and $\overline{\sigma_{C,i}}\,$, respectively. 
As the number of samples increases, 
Eq.~(\ref{update_2mod}) returns to Eq.~(\ref{update_2})
\footnote{
When zero divided by zero appears owing to rounding error, we approved this update by setting $z_i=1$ 
in the numerical simulations in section 4. 
}. 

Secondly, we modify the removal condition, Eq.~(\ref{remove_cost}). 
Since node removal irreversibly decreases the representational power of the RBM, 
we carefully verify whether $C_k \leq 0$ is satisfied. 
However, since the logarithmic function in the second term of Eq.~(\ref{remove_cost}) 
drastically decreases in $p(h_k=0) < 1\,$, 
a small sampling error in $p(h_k=0)$ results in a large error in $\ln p(h_k=0)\,$, 
which makes it difficult to evaluate the removal cost accurately by Gibbs sampling. 
Therefore, we employ an upper bound of $C_k$ as an effective removal cost, $C_k'\,$: 
 \begin{eqnarray}
  C_k &=& -\sum_\v q(\v) \, \ln p(h_k = 0 | \v) + \ln [1-p(h_k = 1)] \nonumber \\ 
   &\leq& -\sum_\v q(\v) \, \ln p(h_k = 0 | \v) -p(h_k = 1) \nonumber \\ 
   &\equiv& C'_k \, .
  \label{upper_rc}
 \end{eqnarray}
Then, consider the approximation of $C'_k$ by Gibbs sampling, 
 \begin{eqnarray}
  \overline{C'_k} &\equiv& -\frac{1}{S} \, \sum_{\alpha=1} ^S \ln p(h_k = 0 | \v^\alpha) 
    -\frac{1}{S} \, \sum_{\alpha=1} ^S h_k^\alpha \, , \label{sample_rc}
 \end{eqnarray}
where $\alpha$ is the sample index. 
Since samplings from $q(\v)$ and $p(\v, \h)$ are independent, 
the first and second terms of Eq.~(\ref{sample_rc}) have no correlations. 
Thus, when the sampling size, $S\,$, is sufficiently large, 
the probability distribution of $\overline{C'_k}$ can be approximated 
by the normal distribution, due to the central limit theorem: 
% One finds that approximation of $C'_k$ by Gibbs sampling, $\overline{C'_k}\,$, follows the central limit theorem: 
 \begin{eqnarray}
  \overline{C'_k} 
%   &\equiv& -\frac{1}{S_1} \, \sum_{\alpha=1} ^{S_1} \ln p(h_k = 0 | \v^\alpha) \nonumber \\ 
%   && -\frac{1}{S_2} \, \sum_{\alpha=1} ^{S_2} \delta_{h_k^\alpha , 1} \nonumber \\ 
   &\sim& {\cal N} \left( C'_k, \frac{\sigma_1^2}{S} + \frac{\sigma_2^2}{S} \right) \, ,
  \label{sample_rc_clt} \\
  \sigma_1^2 &=& \sum_\v q(\v) \, \left[ \ln p(h_k=1|\v) \right]^2 
                   - \left[ \sum_\v q(\v) \, \ln p(h_k=1|\v) \right]^2 \, , \\[0.3cm]
  \sigma_2^2 &=& p(h_k=1) - [p(h_k=1)]^2 \, ,
 \end{eqnarray}
where ${\cal N}(\mu,\sigma^2)$ denotes the normal distribution. 
The unbiased standard deviation of $\overline{C'_k}$ is given by 
 \begin{eqnarray}
  \overline{\sigma_{C'_k}} = \sqrt{\frac{\overline{\sigma_1^2} + \overline{\sigma_2^2}}{S}} \, , 
  \label{effective_std} 
 \end{eqnarray}
where $\overline{\sigma_1^2}$ and $\overline{\sigma_2^2}$ are 
the unbiased variances of $\ln p(h_k|\v)$ and $h_k\,$, respectively. 
Using $\overline{C'_k}$ and $\overline{\sigma_{C'_k}}$, 
% \begin{eqnarray}
%  \overline{\sigma_{C'_k}} = \sqrt{\frac{\overline{\sigma_1^2} + \overline{\sigma_2^2}}{S}} \, , 
%  \label{effective_std} 
% \end{eqnarray}
%where $\overline{\sigma_1^2}$ and $\overline{\sigma_2^2}$ are 
%the unbiased variances of $\ln p(h_k|\v)$ and $h_k$ in Eq.~(\ref{sample_rc}), respectively, 
we change the removal criterion from $C_k \leq 0$ 
into $\overline{C'_k} + a\, \overline{\sigma_{C'_k}} \leq 0 \,$, 
where $a$ tunes the confidence intervals of $C'_k\,$. 
By increasing $a\,$, we can decrease the probability 
that a hidden unit is wrongly removed when its true removal cost is positive, $C_k>0\,$. 
When $\overline{\sigma_{C'_k}} / D$ is not small, 
this incorrect removal may harm the performance. 
Thus, a large $a$ is used to decrease the probability of an incorrect removal.

In summary, our node removal procedure is as follows (Alg.~\ref{alg:remove}). 
First, we remove all hidden units that satisfy the modified removal condition. 
Then, at each parameter update step, 
we choose the smallest removal cost and decrease it using Eq.~(\ref{update_2mod}) 
until a hidden unit can be removed. 
The source code is available on GitHub at https://github.com/snsiorssb/RBM.

 \begin{algorithm}[H]
 \caption{Node removal procedure}
 \label{alg:remove}
 \begin{algorithmic}[1]
 \For{number of removing iterations}
  \Repeat
   \State obtain $S$ realizations, $(\v^1,\h^1), \ldots, (\v^S, \h^S)\,$ by $n$-step block Gibbs sampling (PCD-$n$). 
   \State evaluate $\overline{C'_j}$ for all remaining hidden units by using Eq.~(\ref{sample_rc}). 
   \State determine a node to be removed, $k = \argmin_j \overline{C'_j}\,$. 
   \State evaluate $\overline{\sigma_{C'_k}}$ by using Eq.~(\ref{effective_std}). 
   \If {$\overline{C'_k} + a\, \overline{\sigma_{C'_k}} \leq 0$} 
   \State remove the target node 
   \State obtain $S$ realizations by Gibbs sampling (tempered transition, from $\beta_0=1$ to $\beta_1$ divided by $l$ intervals).
   \EndIf
  \Until{$\overline{C'_j} + a\, \overline{\sigma_{C'_j}} > 0$ for any $j\,$.}
  \State evaluate $\overline{\partial_i D}\,$, $\overline{\partial_i C_k}\,$, $\overline{\sigma_{D,i}}\,$, 
  and $\overline{\sigma_{C,i}}\,$. 
  \State determine $\overline{\Delta\bxi}$ from Eq.~(\ref{update_2mod}). 
  \State $\bxi^{s+1} = \bxi^s - \nu \, \overline{\Delta\bxi} \, $. 
 \EndFor
 % \Until{$|\Delta\bxi|^2 < \theta\,$.} 
 % \Until{$|\Delta\bxi|^2 = 0\,$.}
 \end{algorithmic}
 \end{algorithm}

 \section{Numerical simulation}
\begin{figure}[t]
%\begin{figure}[t]
 \begin{center}
  \includegraphics[width=60mm]{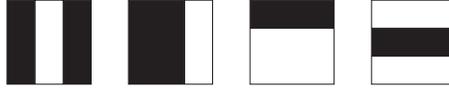}
 \end{center}
 \caption{Examples of $3\times 3$ Bars-and-Stripes images are shown, 
which are generated as follows. 
First, a white square of $A \times A$ pixel is prepared. 
Next, each column of the square is painted black with probability $1/2\,$. 
Finally, the square is rotated $90^\circ$ with probability $1/2\,$. 
For $A=3\,$, $14$ different images are created. \vspace{0.7cm}
}
 \label{BS_samp}
\end{figure}

In this section, we show that the proposed algorithm does not spoil the performance of the RBMs 
by using two different datasets. 
First, we used the $3 \times 3$ Bars-and-Stripes dataset (\cite{mackay2003information}) (Fig.~\ref{BS_samp}), 
which is small enough to allow calculation of the exact KLD during the removal processes. 
%First, we used $3 \times 3$ Bars-and-Stripes 
%and monitored the change of the KLD, which is exactly calculated, through removal processes. 
Next, we employed MNIST dataset of handwritten images (\cite{lecun1998mnist}) 
and verified that our algorithm also works in realistic-size RBMs. 
% We first train the RBM, and then, we perform removal process twice starting from the same trained RBMs. 

Since parameter update after sufficient learning slightly changes $p(\v, \h)\,$, 
it can be considered that short Markov chains are enough for convergence to $p(\v, \h)$ after parameter updates. 
Thus, we used PCD (\cite{tieleman2008training}) with $n$-step block Gibbs sampling (PCD-$n$) 
in both learning and removal processes, except for samplings immediately after a node removal. 
However, a change of $p(\v, \h)$ caused by node removal is expected to be larger 
than that caused by parameter updates. 
Hence, PCD-$n$ with small $n$ may not converge to $p(\v, \h)$ 
and may fail to sample from $p(\v, \h)$ immediately after node removals. 
Thus, we carefully performed Gibbs sampling using tempered transition (\cite{salakhutdinov2009learning, neal1996sampling}) 
at these times. 
In tempered transition, we linearly divided the inverse temperature 
from $\beta_0=1$ to $\beta_1=0.9$ into $l=100$ intervals. 
% following Ref.~\cite{salakhutdinov2009learning}. 
% We note here that since maximization of the performance is not seriously considered in this paper, 
We did not use a validation set for early stopping or hyperparameter searches 
in both the learning and removal processes. 
% We naively determined them by hand without using a validation set. 

%First, we train the RBM 
%by using PCD-1 as Gibbs sampling with $1,000$ mini-batches and fixed learning rate $\lambda=0.01\,$. 
%Then, we remove hidden units by using PCD-1 as Gibbs sampling 
%with $1,000$ mini-batch and fixed change rate $\nu=0.01\,$. 

  \subsection{Bars-and-Stripes}
%In order to verify whether our algorithm preserves performance of RBMs, 
%we employed $3\times 3$ Bars-and-Stripes \cite{mackay2003information}, 
%where the KLD can be exactly calculated. 
An artificial dataset called Bars-and-Stripes was used 
to demonstrate that our algorithm effectively works when the data distribution is completely known. 
Thus, we did not divide the dataset into training and test sets. 
First, we trained the RBM with $M=9$ visible units and $N=30$ hidden units 
using PCD-5 and PCD-1 with a batch size of $100$ and a fixed learning rate, $\lambda = 10^{-2}\,$. 
After $50,000$ learning steps, we performed removal processes starting from the same trained RBM 
with a batch size of $1,000$ and a fixed parameter change rate, $\nu = 10^{-2}\,$. 
During the beginning of the removal process, 
the typical value of $\overline{\sigma_{C'_k}} / D$ was not small, that is, $\overline{\sigma_{C'_k}} / D \sim 0.1\,$. 
Thus, we employed a strict removal criterion, $\overline{C_k'} + 3\, \overline{\sigma_{C'_k}} \leq 0\,$. 
%In this simulation, the number of visible units is $9\,$, 
%and therefore, we can exactly evaluate the KLD, $\DKL{q}{p}\,$. 
%We demonstrate two numerical simulations by using the same initial condition, 
%As a node removal criterion, we employ $C'_k + 3\sigma_{C_k} \leq 0 \,$, 
%where $\sigma_{C_k}$ is standard deviation of $C'_k\,$. 
%That is, $a=3,\, \theta_{\rm cut}=0$ in Alg.~\ref{alg:remove}. 
%Immediately after we remove a hidden unit, 
%we employ tempered transition \cite{salakhutdinov2009learning} in order to safely sample from $p(\v, \h)\,$. 

\begin{figure}[t]
 \begin{center}
  \includegraphics[width=90mm]{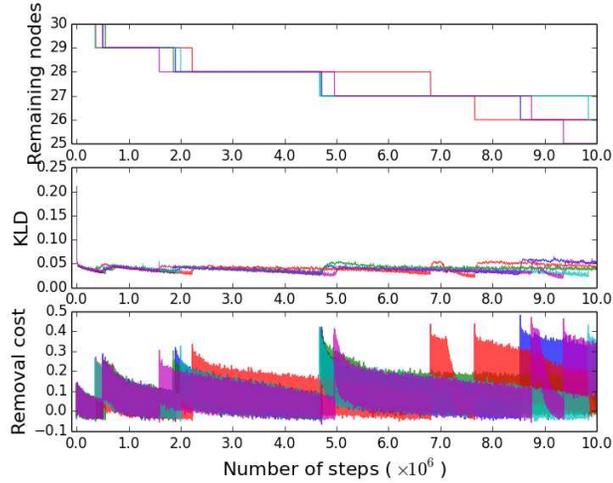}
 \end{center}
 \caption{The number of hidden units $N$ (top), KLD (middle), 
and smallest removal cost (bottom) are shown 
as functions of the number of removal steps. 
The $3\times 3$ Bars-and-Stripes dataset was employed. 
PCD-5 was used for block Gibbs sampling. 
Each color corresponds to a different trial. \vspace{0.7cm}
}
 \label{fig:BS_new}
\end{figure}

\begin{figure}[t]
 \begin{center}
  \includegraphics[width=90mm]{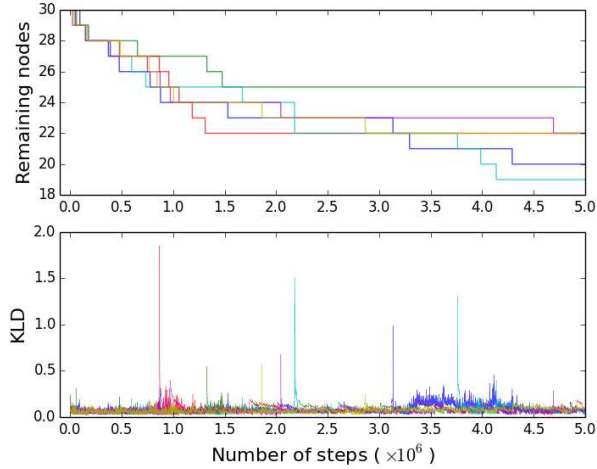}
 \end{center}
 \caption{The number of hidden units $N$ (top),  and KLD (bottom) are shown 
as functions of the number of removal steps. 
The $3\times 3$ Bars-and-Stripes dataset was employed. 
PCD-1 was used for block Gibbs sampling. 
Each color corresponds to a different trial. \vspace{0.7cm}
}
 \label{fig:BS}
\end{figure}

\begin{figure}[t]
 \begin{center}
  \includegraphics[width=90mm]{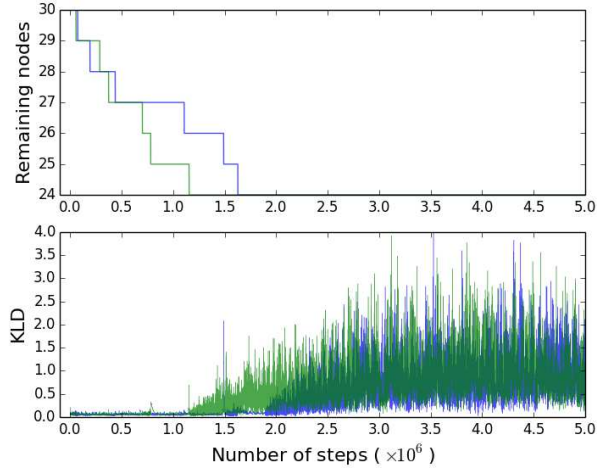}
 \end{center}
 \caption{Two trials failed to keep the KLD in case of the PCD-1. 
Large fluctuations of the KLD appear 
immediately (green) and sufficiently (blue) after node removal. \vspace{0.7cm}
}
 \label{fig:BS_app}
\end{figure}

The results are shown in Figures~\ref{fig:BS_new}, \ref{fig:BS}, and \ref{fig:BS_app}. 
We stopped the removal processes after $10,000,000$ steps in Figure~\ref{fig:BS_new} 
and after $5,000,000$ steps in Figures~\ref{fig:BS} and \ref{fig:BS_app}. 
The removal procedure employing PCD-5 slowly decreases $N$ with small fluctuations of the KLD 
in all five trials (Figure~\ref{fig:BS_new}). 
In particular, the removal cost in Figure~\ref{fig:BS_new} shows that 
if a hidden unit with the smallest removal cost is removed before it decreases, 
then the KLD increases approximately sevenfold. 
This result clearly shows that the update rule, Eq.~(\ref{update_2mod}), 
is useful for maintaining the performance during the removal processes. 
The removal procedure employing PCD-1 decreases $N$ more rapidly while approximately preserving the KLD 
in six out of eight trials (Figure~\ref{fig:BS}), 
although some sharp peaks appear in the change of the KLD after node removals. 
However, two out of eight trials that employed PCD-1 fail to preserve the KLD (Figure~\ref{fig:BS_app}).

\begin{figure}[htbp]
%\begin{figure}[t]
\centering
% \begin{center}
  \includegraphics[width=90mm]{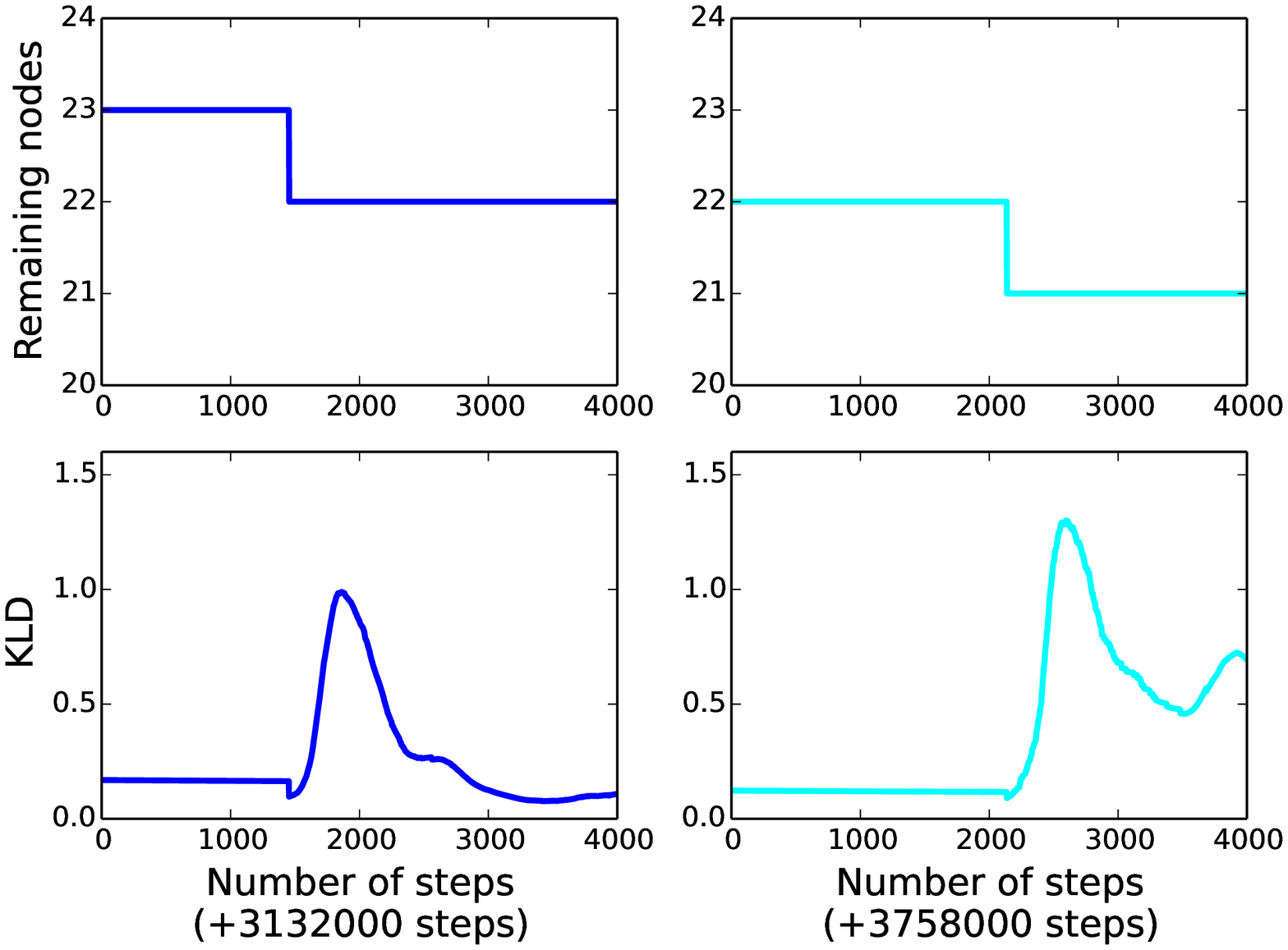}
% \end{center}
 \caption{The peaks after the $3,000,000$th step (blue line) and before the $4,000,000$th step (cyan line) 
in Figure~\ref{fig:BS} are enlarged. 
These figures show that node removal slightly decreases the KLD, 
and parameter updates immediately following removal caused increases in the KLD. \vspace{0.7cm}
%increases of the KLD was caused by parameter update after a node removal 
%rather than removal itself.
}
 \label{fig:large}
\end{figure}

First, we discuss the sharp peaks observed in Figure~\ref{fig:BS}, 
%\sout{One finds that $N$ decreases while approximately keeping the KLD, 
%and the number of the remaining nodes at $5,000,000$th 
%step, when removal processes were stopped, 
%are different among trials, due to stochasticity of the algorithm. 
%In the upper panel of Fig.~\ref{fig:BS}, which shows the behavior of the KLD, 
%some peaks appear after node removals.} 
%In this algorithm, 
which resulted from inaccurate estimates of $C'_k$ or $\overline{\Delta\bxi}\,$. 
In order to distinguish among them, 
we enlarge peaks in the change of the KLD (Figure~\ref{fig:large}) 
and find that these peaks were caused by the failure of Gibbs sampling in 
parameter updates immediately after node removals rather than node removals themselves. 
This behavior supports the assumption that the change of $p(\v,\h)$ caused by node removal can be large 
and 
% than that caused by parameter updates, 
%short Markov chains employed in CD and PCD may not reach $p(\v,\h)$ after one-step parameter updates. 
can result in failure of Gibbs sampling. 
Nevertheless, owing to the tempered transition, 
%peaks after node removals were mostly suppressed, 
%although we did not show the result without the tempered transition in this paper. 
most of the parameter updates after node removal produced rather small peaks in Figure~\ref{fig:BS}.

Next, we discuss large fluctuations of the KLD in Figure~\ref{fig:BS_app}. 
Failure of Gibbs sampling through parameter updates is expected to occur more frequently 
as the removal process continues for the same reason as in the learning process 
(\cite{fischer2010empirical, desjardins2010tempered}). 
It can be considered that the problem in the learning process arises as follows. 
At the beginning of the learning process, the RBM parameters are approximately zero, 
and $p(\v)$ is almost a uniform distribution. 
As the leaning proceeds, each component of $\bxi$ is expected to move away from zero 
in order to adjust $p(\v)$ to the data distribution, $q(\v)\,$. 
%\sout{In the learning process, absolute values of $\xi_i$'s are expected to become large, 
%in order to adjust $p(\v)\,$, which is an almost uniform distribution at the beginning of training, to $q(\v)\,$. }
In the removal process, components of $\bxi$ are also expected to move away from zero 
in order that the remaining system compensates for the roles of the removed hidden units. 
As one can find from Eq.~(\ref{propup}) and Eq.~(\ref{propdown}), 
the transition matrices used in MCMC, $p(\h|\v)$ and $p(\v|\h)\,$, take almost either $0$ or $1$ 
in the region where $|\bxi|$ is large. 
Therefore, block Gibbs sampling behaves almost deterministically. 
Hence, dependence on the initial condition remains for a long time, or equivalently, 
it takes a long time to converge to $p(\v, \h)$ even after a one-step parameter update in the large $|\bxi|$ region. 
Thus, the model distribution after parameter update, from which we should sample, may be quite different from 
the probability distribution after a few block Gibbs sampling steps. 
As a result, parameters are updated using inaccurate Gibbs samples. 
If these deviations are corrected by subsequent parameter updates, then the KLD decreases again. 
However, if the failure of Gibbs sampling continues for a long time, 
then the KLD drastically fluctuates. 
From Figure~\ref{fig:BS_app}, it can be found that 
such a drastic increase in the KLD can emerge not only immediately after node removal (green line) 
but also later (blue line). 
Therefore, in order to prevent the problem resulting from a long convergence time of the block Gibbs sampling, 
% \sout{Markov Chain Monte Carlo (MCMC), }
the removal process should be stopped at some point in time as with the learning process.

  \subsection{MNIST}
%In order to verify whether the proposed algorithm also preserves performance of RBMs with a large number of nodes, 
%we trained an RBM on the MNIST dataset, which requires the RBM to have at least 784 visible nodes. 
We used $60,000$ out of $70,000$ MNIST images 
for the evaluation of $C_k\,$, $\nabla_\bxi C_k\,$, and $\nabla_\bxi D$ 
in the learning and removal processes. 
Each pixel value was probabilistically set to $1$ proportional to its intensity 
(\cite{salakhutdinov2008quantitative, tieleman2008training}). 
We first trained the RBM with $M=784$ visible units and $N=500$ hidden units 
using PCD-1 with a batch size of $1,000$ and fixed learning rate $\lambda=10^{-2}\,$. 
After $200,000$ learning steps, we performed the removal processes starting from the same trained RBM 
with a batch size of $1,000$ and a fixed parameter change rate, $\nu = 10^{-2}\,$. 
In this case, the typical value of $\overline{\sigma_{C'_k}} / D$ at the first removal step is small, 
that is, $\overline{\sigma_{C'_k}} / D \sim 10^{-4}\,$. 
Thus, we employed $\overline{C_k'} + \overline{\sigma_{C'_k}} \leq 0$ as the removal criterion 
in order to quickly remove hidden units under the restriction that they do not drastically decrease the performance.

As mentioned in section 2, the KLD cannot be evaluated, 
owing to unknown probability $q(\v)$ and a large state space of the RBM. 
%\com{($R$ uses training set and is the evaluation criterion. 
%$\wt D$ requires all data and is just measure that shows effectiveness of our method.)} 
Thus, we employed an alternative evaluation criterion, namely, 
the KLD of $p(\v)$ from empirical distribution of samples generated from the test set, $q_{\rm d}(\v)\,$, 
 \begin{eqnarray}
  \wt D 
   &\equiv& \DKL{q_{\rm d}}{p} \nonumber \\ 
   &=& \sum_\v q_{\rm d}(\v) \, \ln q_{\rm d}(\v) +\ln Z \nonumber \\
   && + \sum_\v q_{\rm d}(\v) \, \left[\sum_i b_i\, v_i + \sum_j \ln \left( 1 + {\rm e}^{c_j + \sum_i v_i \, w_{ij}} \right) \right]\, , 
 \end{eqnarray}
where $Z$ is the normalization constant of $p(\v)$ 
and was evaluated by annealed importance sampling (AIS) 
(\cite{neal2001annealed}). 
In the AIS, we used $100$ samples and 
linearly divided the inverse temperature from $\beta = 0$ to $\beta = 1$ into $10,000$ intervals. 
%However, it takes quite a long time to evaluate $Z$ by AIS. 
%Therefore, we calculated $\wt D$ at every $5,000$ steps 
%and employed another evaluation criterion, the reconstruction error, for reference, 
%between the intervals of evaluations of $\wt D\,$. 
Since the evaluation of $Z$ by AIS takes a long time, 
% we could not show the change of $\wt D$ at every steps of the removal process 
we calculated $\wt D$ at every $50,000$ step. 
Between the intervals of evaluations of $\wt D\,$, 
we employed another evaluation criterion, the reconstruction error, for reference. 
The reconstruction error, $R\,$, can be easily calculated 
and is widely used to roughly estimate the performance of the RBM 
(\cite{bengio2007greedy, taylor2007modeling, hinton2012practical}): 
%Thus, we employ reconstruction error, $R\,$, 
%which is widely used to measure the performance of the RBM: 
%\com{(Correct equation???)}
 \begin{eqnarray}
  R &=& -\frac{1}{S} \, \sum_{\alpha=1}^S \sum_{i=1}^M \left[ v_i^\alpha \, \ln \wt v_i^\alpha 
     + (1-v_i^\alpha) \, \ln (1-\wt v_i^\alpha) \right] \, , \\[0.3cm]
  \wt v_i^\alpha 
   &=& \frac{ {\rm e}^{b_i + \sum_j w_{ij} \wt h_j^\alpha } }{ 1 + {\rm e}^{b_i + \sum_j w_{ij} \wt h_j^\alpha} } \, , \\[0.3cm]
  \wt h_j^\alpha 
   &=& \frac{ {\rm e}^{c_j + \sum_i v_i^\alpha w_{ij}} }{ 1 + {\rm e}^{c_j + \sum_i v_i^\alpha w_{ij}} } \, ,
 \end{eqnarray}
where $\alpha$ denotes the index of a mini-batch, and $\v^\alpha$ is a sample from the training set. 
%\sout{$q_{\rm d}(\v)\,$. 
%On one hand, $\wt D$ evaluated by AIS is considered to reflect the performance of RBMs with high precisions 
%and takes a long time for calculation. 
%On the other hand, $R$ is not considered to reflect the change of the KLD precisely 
%and takes short time for calculation. 
%Thus, in order to show the change of the performance, 
%we employed $\wt D$ every $5,000$ steps and used $R$ during removal processes for reference. 
%}
% we mainly employed $\wt D$ as an evaluation criterion and used $R$ for reference. 

\begin{figure}[H]
 \begin{center}
  \includegraphics[width=150mm]{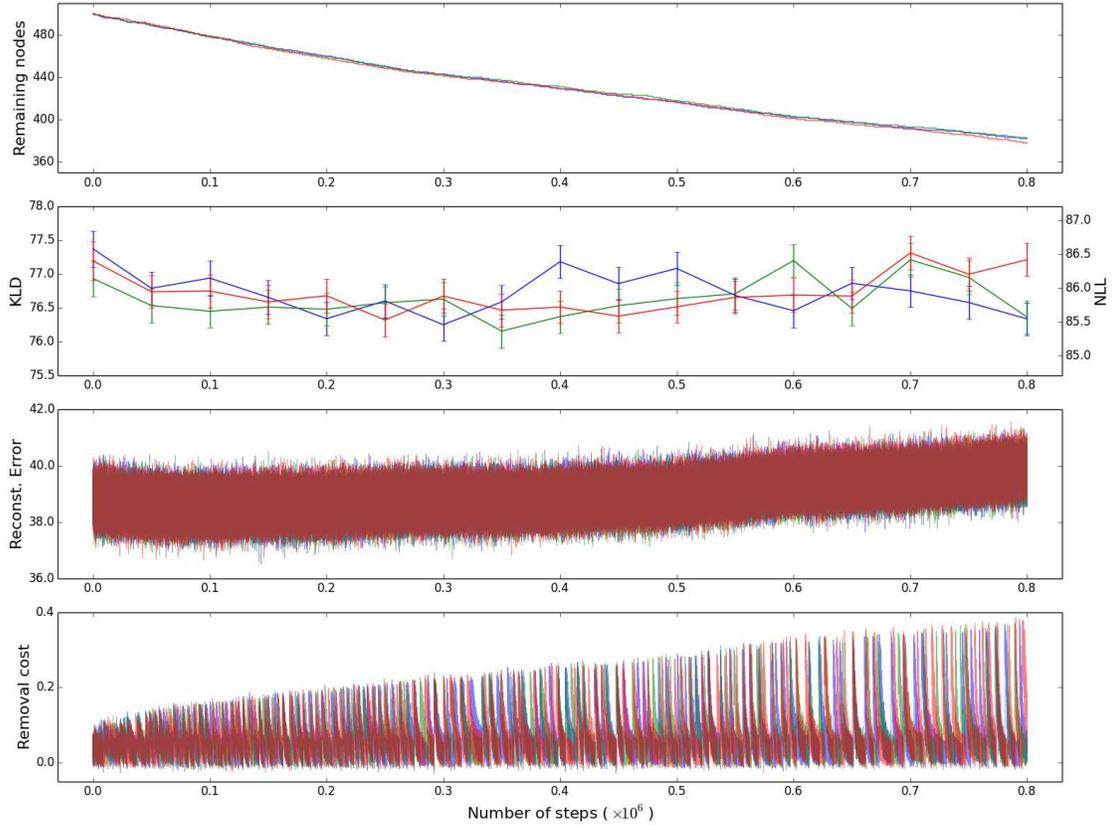}
 \end{center}
 \caption{From the top to the bottom, the number of hidden units $N\,$, 
the KLD of $p(\v)$ from $q_{\rm d}(\v)\,$, the reconstruction error $R\,$, and the effective removal cost 
are shown as functions of the number of removal steps. 
MNIST handwritten images were employed as the dataset. 
Each color corresponds to a different trial. 
In the second panel from the top, the width of the KLD represents $1\sigma$ confidence intervals, 
and the negative log-likelihood (NLL), 
$-l \equiv \wt D -\sum_\v q_{\rm d}(\v) \, \ln q_{\rm d}(\v)\,$, 
is also shown for the evaluation of the performance together with $\wt D\,$. 
\vspace{0.7cm}
}
 \label{fig:MNIST}
\end{figure}

\begin{figure}[htbp]
 \begin{center}
  \includegraphics[width=100mm]{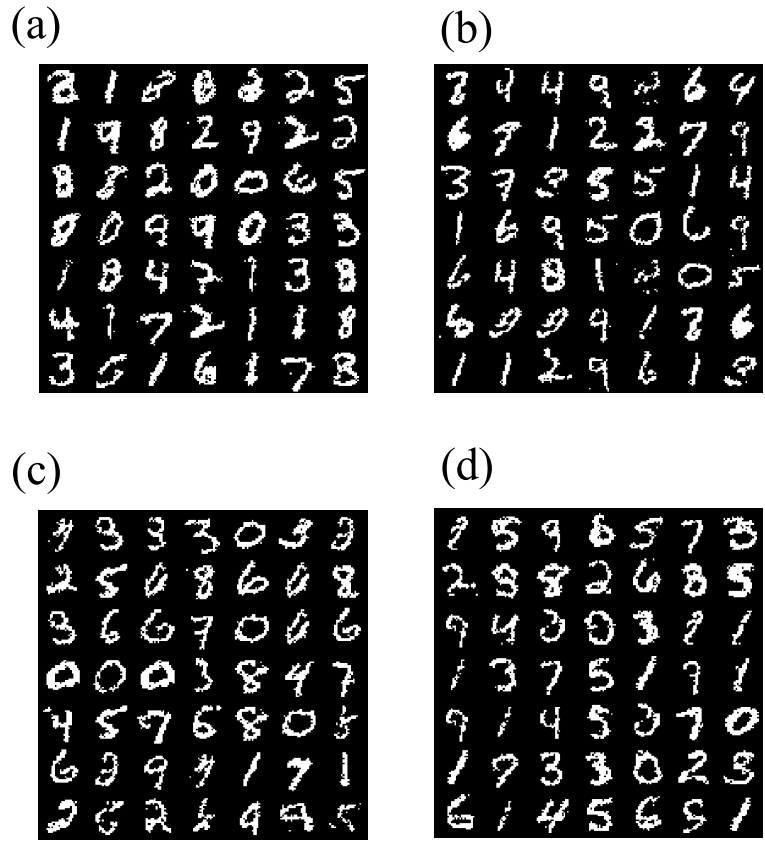}
 \end{center}
 \caption{MNIST images are shown at the start and ends of the removal processes. 
(a) Samples of visible configurations at the $0$th step of the removal processes. 
(b, c, d) Samples of visible configurations 
at the $800,000$th step of the blue, green, and red lines  in Figure~\ref{fig:MNIST}, respectively. 
\vspace{0.7cm}
}
 \label{fig:images}
\end{figure}

The progress of the removal processes is shown in Figure~\ref{fig:MNIST}, 
and samples of visible variables at the beginning and the end of the removal processes are presented in Figure~\ref{fig:images}. 
From the behavior of $N\,$, $\wt D\,$, and $R$ in Figure~\ref{fig:MNIST}, 
it can be found that in a realistic-size RBM, 
our algorithm decreases the number of hidden units while avoiding a drastic increase in the KLD
\footnote{
Fig. 7 shows that the increase of the reconstruction error does not mean the increase of the KLD. 
However, it may be used as a stopping criterion which can be easily calculated. 
}. 
% \sout{effectively works. }
We stopped three removal processes after $800,000$ steps, 
and the RBMs were compressed to $N\sim 400\,$. 
The number of removal steps is much larger than that of the learning steps. 
However, this is not a defect of our algorithm, 
since our motivation is not to quickly compress the RBM but to preserve its performance during the removal process. 
As a reference for the performance of the compressed RBMs, 
% \sout{after the removal process}
we trained the RBM with $N=400$ using the same setting 
employed in the learning of the RBM with $N=500\,$. 
The performance of this RBM was $\wt D = 78.0 \pm 0.3$ 
(where $\pm$ indicates $1\sigma$ confidence interval), 
which is almost the same performance of the RBMs after the removal process. 
This result suggests that our algorithm does not harm the performance, 
although we did not highly optimize the learning process for the RBMs with $N=400$ and $N=500\,$. 
The gradual increase of the upper side of $C'_k$ in Figure~\ref{fig:MNIST} supports our intuitive explanation 
that the contribution of the remaining hidden units to the performance increases 
in order to maintain the performance. 
%\sout{Behaviors of $N$ and $\wt D$ in Figure~\ref{fig:MNIST} are qualitatively similar to 
%those of $N$ and the KLD in Figures~\ref{fig:BS_new} and \ref{fig:BS}, 
%and hence, interpretation of the results is the same as the numerical simulation in the previous subsection. }
%However, since large absolute values of RBM parameters slow convergence of Markov chain Monte Carlo, 
%which is used in block Gibbs sampling \cite{fischer2010empirical, desjardins2010tempered}, 
%removing process does not always preserve performance. 
Thus, also in this case, 
an extremely long removal process can increase $|\bxi|$ and may lead to failure of Gibbs sampling. 
% \sout{which may successively increase the KLD. }
Thus, the removal process should be stopped before a successive increase in the KLD occurs. 
% \sout{it spoils the performance. }
Since the KLD cannot be evaluated in large-size RBMs, 
we recommend monitoring the change in performance 
by employing some evaluation criterion used in the learning process in previous studies, 
e.g. the reconstruction error (\cite{bengio2007greedy, taylor2007modeling, hinton2012practical}), 
the product of the two probabilities ratio (\cite{buchaca2013stopping}), 
and the likelihood of a validation set obtained by tracking the partition functions (\cite{desjardins2011tracking})
\footnote{
Tracking the partition function requires the parallel tempering for Gibbs sampling 
%\sout{For tracking the partition function, 
%one has to employ the parallel tempering for Gibbs sampling throughout the numerical simulation }
instead of CD or PCD. 
}
.

%\begin{figure}[t]

 \section{Summary and discussion} 
In this paper, we aimed to decrease the number of hidden units of the RBM without affecting its performance. 
For this purpose, we have introduced the removal cost of a hidden unit 
and have proposed a method to remove it while avoiding a drastic increase in the KLD. 
Then, we have applied the proposed method to two different datasets 
and have shown that the KLD was approximately maintained during the removal processes. 
The increase in the KLD observed in the numerical simulations was caused by the failure of Gibbs sampling, 
which is also a problem in the learning process. 
The RBM has been facing difficulties such as 
accurately obtaining expectation values that are computationally expensive. 
Several kinds of Gibbs sampling methods have been proposed 
(\cite{hinton2002training, tieleman2008training, tieleman2009using, salakhutdinov2009learning, cho2010parallel, 
 desjardins2010tempered}), 
which provide precise estimates and increase the performance of the RBM. 
However, more accurate Gibbs sampling methods require a longer time for evaluations. 
If expectation values can be precisely evaluated, then our algorithm is expected be more effective. 
We expect that physical implementation of the RBM (\cite{dumoulin2014challenges}) 
% \cite{ly2009high, dumoulin2014challenges} 
becomes an accurate and fast method for their evaluation. 
%\sout{of the expectation values 
%with respect to the probability distribution of an RBM. }
%removal cost given by Eq.~(\ref{remove_cost}) can be applied to BM and DBM, 
%although its evaluation takes long time. 

Finally, we comment on another application of the removal cost. 
If the representational power of the system is sufficient, 
then an arbitrary hidden unit can be safely removed by decreasing its removal cost. 
Hence, by repeatedly adding and removing hidden units, 
entire hidden units of a system can be replaced. 
Such a procedure may be useful for reforming physically implemented systems 
that are difficult to copy and must not be halted. 
%For future works, we deal with learning systems which have spatial and temporal structures 
%and consider how to change network structure without changing its performance. 

 \section{Acknowledgment}
This research is supported by JSPS KAKENHI Grant Number 15H00800. 

\appendix

 \section{Derivation of Eq.~(\ref{remove_cost})}
For convenience, we introduce two unnormalized probabilities, 
$p^*(\v, \h) = {\rm e}^{-E(\v, \h)}$ and 
$p^*_{\backslash k}(\v, \h_{\backslash k}) = {\rm e}^{-E(\v,\h)} |_{h_k = 0} \, $. 
Then, we can obtain $C_k$ as follows: 
 \begin{eqnarray}
  C_k &=& \DKL{q}{p_{\backslash k}} - \DKL{q}{p} \nonumber \\ 
   &=& \sum_\v q(\v) \, \ln \frac{q(\v)}{\sum_{\h_{\backslash k}} p_{\backslash k}(\v, \h_{\backslash k})}
    - \sum_\v q(\v) \, \ln \frac{q(\v)}{\sum_\h p(\v, \h)} \nonumber \\[0.3cm]
   &=& -\sum_\v q(\v) \, \ln \sum_{\h_{\backslash k}} p_{\backslash k}(\v, \h_{\backslash k}) 
      + \sum_\v q(\v) \, \ln \sum_\h p(\v, \h) \nonumber \\[0.3cm] 
   &=& -\sum_\v q(\v) \, \ln \frac{ \sum_{\h_{\backslash k}} p_{\backslash k}^*(\v, \h_{\backslash k}) }
                                              { \sum_{\v', \h'_{\backslash k}} p_{\backslash k}^*(\v', \h'_{\backslash k}) } 
      + \sum_\v q(\v) \, \ln \frac{ \sum_\h p^*(\v, \h) }
                                                { \sum_{\v', \h'} p^*(\v', \h') } \nonumber \\[0.3cm] 
   &=& -\sum_\v q(\v) \, \ln \frac{ \sum_{\h_{\backslash k}} p^*(\v, \h)|_{h_k=0} }
                                              { \sum_{\v', \h'_{\backslash k}} p^*(\v', \h')|_{h_k=0} } 
      + \sum_\v q(\v) \, \ln \frac{ \sum_\h p^*(\v, \h) }
                                                { \sum_{\v', \h'} p^*(\v', \h') } \nonumber \\[0.3cm] 
   &=& -\sum_\v q(\v) \, \ln \frac{\sum_{\h_{\backslash k}} p^*(\v, \h)|_{h_k=0}}
                                              {\sum_\h p^*(\v, \h)} 
    + \ln \frac{\sum_{\v', \h'_{\backslash k}} p^*(\v', \h')|_{h_k=0}}
                       {\sum_{\v', \h'} p^*(\v', \h')} \nonumber \\[0.3cm]
   &=& -\sum_\v q(\v) \, \ln p(h_k=0 | \v) + \ln p(h_k=0) \, .
%   &=&  -\sum_\v q(\v) \, \ln \left[ 1 - p(h_k=1 | \v) \right] \nonumber \\
%   && + \ln \left[1 -  \sum_{\v, \h} h_k \, p(\v, \h) \right] \, .
 \end{eqnarray}

Next, consider the simultaneous removal of several hidden units. 
Suppose $\tra\k=(k_1, \ldots, k_r)$ denotes the indices of the hidden units to be removed 
and define $p_{\backslash \k}(\v)$ as the probability distribution after removal of these hidden units. 
Following a similar calculation above, we obtain the removal cost for several hidden units: 
 \begin{eqnarray}
  C_\k &\equiv& \DKL{q}{p_{\backslash \k}} - \DKL{q}{p} \nonumber \\ 
   &=& -\sum_\v q(\v) \, \ln p(h_{k_1} = \cdots = h_{k_r} = 0 | \v) \nonumber \\
   && + \ln p(h_{k_1} = \cdots = h_{k_r} = 0) \, .
  \label{multi_rc}
 \end{eqnarray}
In the case of the RBM, the first term of Eq.~(\ref{multi_rc}) can be simplified as 
 \begin{eqnarray}
  C_\k &=& \DKL{q}{p_{\backslash \k}} - \DKL{q}{p} \nonumber \\ 
   &=& -\sum_\v q(\v) \, \sum_{\alpha=1}^r \ln p(h_{k_\alpha} = 0 | \v) \nonumber \\
   && + \ln p(h_{k_1} = \cdots = h_{k_r} = 0) \, .
  \label{multi_rc_RBM}
 \end{eqnarray}
This removal cost can be used to minimize the size of the RBM. 
Suppose $D_0$ is the KLD to be preserved. 
If some set of parameters, $\bxi\,$, satisfies $C_\k =0$ and $D=D_0\,$, 
then the hidden units whose indices are $\k$ can be removed simultaneously without changing the KLD. 
Furthermore, if one can find a set $\bxi$ that can remove as many hidden units as as possible, 
then the size of the RBM is minimized. 
However, finding such a set of parameters is difficult problem.

 \section{Change of $D$ and $C_k$ by the naive update rule, Eq.~(\ref{update_2})}
In this Appendix, we show that the naive update rule, Eq.~(\ref{update_2}), 
decreases both $D$ and $C_k$ at ${\cal O}(|\Delta\bxi|)\,$. 
The change of $D$ and $C_k$ by Eq.~(\ref{update_2}) at ${\cal O}(|\Delta\bxi|)$ are given by 
 \begin{eqnarray}
  \frac{\partial D}{\partial \xi_i} \, \Delta\xi_i 
   &=& -\nu \cdot \theta \left( \frac{\partial D}{\partial \xi_i} \, \frac{\partial C_k}{\partial \xi_i} \right)
   \cdot \left( \frac{\partial D}{\partial \xi_i} \right)^2 \, , \label{D_1st} \\[0.3cm]
  \frac{\partial C_k}{\partial \xi_i} \, \Delta\xi_i 
   &=& -\nu \cdot \theta \left( \frac{\partial D}{\partial \xi_i} \, \frac{\partial C_k}{\partial \xi_i} \right)  
   \cdot \frac{\partial D}{\partial \xi_i} \cdot \frac{\partial C_k}{\partial \xi_i} \, . \label{C_1st} 
 \end{eqnarray}
In the case of $\partial D/\partial\xi_i \cdot \partial C_k/\partial\xi_i \geq 0\,$, 
Eq.~(\ref{D_1st}) and Eq.~(\ref{C_1st}) become
 \begin{eqnarray}
  \frac{\partial D}{\partial \xi_i} \, \Delta\xi_i 
   &=& -\nu \cdot \left( \frac{\partial D}{\partial \xi_i} \right)^2 \leq 0 \, , \\[0.3cm]
  \frac{\partial C_k}{\partial \xi_i} \, \Delta\xi_i 
   &=& -\nu \cdot \frac{\partial D}{\partial \xi_i} \cdot \frac{\partial C_k}{\partial \xi_i} \leq 0 \, , 
 \end{eqnarray}
and in the case of $\partial D/\partial\xi_i \cdot \partial C_k/\partial\xi_i < 0\,$, 
Eq.~(\ref{D_1st}) and Eq.~(\ref{C_1st}) become 
 \begin{eqnarray}
  \frac{\partial D}{\partial \xi_i} \, \Delta\xi_i 
   &=& 0 \, , \\[0.3cm]
  \frac{\partial C_k}{\partial \xi_i} \, \Delta\xi_i 
   &=& 0 \, . 
 \end{eqnarray}
In both cases, Eq.~(\ref{D_1st}) and Eq.~(\ref{C_1st}) take non-positive values. 
Thus, this update rule decreases both $D$ and $C_k$ at ${\cal O}(|\Delta\bxi|)\,$.

\begin{comment}
 \section{Failure cases of removal procedure in Bars-and-Stripes}
\com{(Move to Sec.4A.)}
In this Appendix, we show a breakdown of the removal process by using Bars-and-Stripes as a dataset. 
We employed 
\changed{
almost 
}
the same setting in Sec.4A 
\changed{
except for changing block Gibbs sampling from PCD-5 to PCD-1. 
}
\sout{and obtained two failure cases in eight trials.} 
Fig.~\ref{fig:BS_app} shows that the KLD began to oscillate owing to inaccurate Gibbs sampling estimates 
and that drastic decrease can emerge not only immediately after node removal (green line) 
but also sufficiently after removal (blue line). 
Therefore, in practice, some evaluation criterion of the KLD used in the learning process should be employed 
to monitor the change of the performance.

 \section{Addition of a new node}
Substituting $w_{ik}=0$ into Eq.~(\ref{remove_cost}), one finds that $C_k$ becomes $0\,$, 
 \begin{eqnarray}
  C_k 
   &=& -\sum_\v q(\v) \, \ln \left[ 1 - \frac{1}{1 + {\rm e}^{c_k} } \right] \nonumber \\ 
   && + \ln \left[ 1 - \frac{1}{1 + {\rm e}^{c_k} } \right] \nonumber \\
   &=& 0 \, .
 \end{eqnarray}
Thus, addition of such a new node does not harm system's performance. 
\end{comment}

%%%%%%%%%%%%%%%%%%%%%%%%%%%%%%%%%%%%%%%%%%%%%%%%%%%%%%%%%%%%%%%%%%%%%%%%%%%%%%%%%%%%%%%%%%%%%%%%%%%%%%%%%%%%%%%%%%%%%%%%%%%%%%%%%%%%%%%%

\end{document}